\documentclass[conference]{IEEEtran}
\IEEEoverridecommandlockouts
\usepackage{cite}
\usepackage{amsmath,amssymb,amsfonts}
\usepackage{algorithmic}
\usepackage{graphicx}
\usepackage{graphics}
\usepackage{textcomp}
\usepackage{xcolor}
\usepackage{hyperref}
\usepackage{booktabs}

\usepackage[noabbrev,capitalise]{cleveref}

\def\BibTeX{{\rm B\kern-.05em{\sc i\kern-.025em b}\kern-.08em
    T\kern-.1667em\lower.7ex\hbox{E}\kern-.125emX}}

\begin{document}

\title{Federated Learning for Autoencoder-based Condition Monitoring in the Industrial Internet of Things}

\author{\IEEEauthorblockN{Soeren Becker, Kevin Styp-Rekowski,  Oliver Vincent Leon Stoll, Odej Kao}
    \IEEEauthorblockA{\textit{Technische Universit\"at Berlin}\\
        \textit{Distributed and Operating Systems Research Group} \\
        \{soeren.becker, styp-rekowski, o.stoll, odej.kao\}@tu-berlin.de}
}

\maketitle

\begin{abstract}
Enabled by the increasing availability of sensor data monitored from production machinery, condition monitoring and predictive maintenance methods are key 
pillars for an efficient and robust manufacturing production cycle in the Industrial Internet of Things. The employment of machine learning models to detect and predict
deteriorating behavior by analyzing a variety of data collected across several industrial environments shows promising results in recent works, yet also often requires transferring
the sensor data to centralized servers located in the cloud. Moreover, although collaborating and sharing knowledge between industry sites yields large benefits, 
especially in the area of condition monitoring, it is often prohibited due to data privacy issues.
To tackle this situation, we propose an Autoencoder-based Federated Learning method utilizing vibration sensor data from rotating machines, that allows for a distributed training on edge devices,
located on-premise and close to the monitored machines. Preserving data privacy and at the same time exonerating possibly unreliable network connections of remote sites, our approach enables 
knowledge transfer across organizational boundaries, without sharing the monitored data. 
We conducted an evaluation utilizing two real-world datasets as well as multiple testbeds and the results indicate that our method enables a competitive performance compared to previous results,
while significantly reducing the resource and network utilization.

\end{abstract}


\section{Introduction}
The amount of distributed devices, sensors and connectivity in the Internet of Things (IoT) is increasing steadily and has a decisive impact on a vast variety of application domains \cite{FerrerB0TK21}. Especially in the area of Industry 4.0, the ever-increasing velocity and volume of data generated by Industrial Internet of Things (IIoT) devices combined with sophisticated analysis jobs can benefit the operational efficiency and manufacturing process \cite{Sari2020} and subsequently enable intelligent industrial operations.

In order to enhance the reliability and allow for a cost effective maintenance, data generated by sensors directly connected to production machinery is often utilized by condition monitoring (CM) strategies to predict failures and identify abnormal behavior \cite{kevinpaper}. CM employs a wide range of sensor components , i.e. vibration and temperature sensors, to monitor system characteristics from different domains, model the system behavior and detect i.e. outliers in the data, indicating a damage or degradation \cite{carden_vibration_2004}.
This is especially important when considering that failures can have a significant and cascading impact on i.e. the manufacturing production cycle or the overall operation, which in turn results in the loss of profit and possibly high maintenance costs. An early detection of a failure can lead to a faster repair and reduced follow-up costs.

Traditionally, the data is sent to external servers or the cloud and subsequently used to run analysis jobs like machine learning models, since plain sensors are not offering enough resources. As industry sites are often located far away from actual cloud data centers, the network capabilities might be limited and for instance affected by high latencies or bad network connectivity in general, resulting in congestion on the network link and delayed results. Furthermore, extending already existing industry environments is often done without introducing sophisticated network infrastructure to the cloud and instead mainly relies on remote and possibly unreliable communication means. Thus, especially for remote locations, that are i.e. only connected via 4G or satellite connections, the transmission of (training) data can also involve high costs. 

Although -- due to the increasing prevalence of more powerful smart devices in edge and IIoT environments -- the processing of data is progressively shifted to edge devices located in or in close proximity to industry sites, the actual training of the models is still often conducted in the cloud \cite{becker_towards_2020,wang_toward_2019}.
Especially in the area of IIoT, the transmission of business-sensitive data to the cloud also results in data privacy issues \cite{m_fusion_2021}, which is why recently distributed machine learning paradigms such as Federated Learning (FL) gain in popularity for the IIoT. In FL, the models are trained locally -- close to the actual data sources --  and then aggregated to a global model without exchanging any training data \cite{fed-google}.
This enables the training of ML models in the organizational boundaries of IIoT sites while still allowing for the integration of knowledge of other faculties.

Therefore, in our work we aim to combine condition monitoring technologies with the FL paradigm in order to enable an on-premise training of models directly in the respective IIoT sites, and an aggregation service located in the cloud, which facilitates an inter-site knowledge sharing.
In our paper, we are focusing on condition monitoring for rotating machines such as bearings or pumps and evaluate the resulting models in terms of condition monitoring and anomaly detection capabilities
on two real-world datasets as well as across different testbeds.

\textit{Contributions}. Summarizing, as  main contribution of this paper we:
\begin{itemize}
    \item Provide a Federated Learning based approach for condition monitoring in the IIoT on the example of rotating machines.
    \item Conduct extensive experiments leveraging two real-world datasets and multiple testbeds.
    \item Evaluate the approach in terms of feasibility, accuracy, resource and network usage and compare it against a state of the art baseline method.  
\end{itemize}

\textit{Outline}. The remainder of this paper is structured as follows:
\cref{section:background} provides background on Condition Monitoring and presents the baseline method. In \cref{section:contribution}, the FL approach is described in detail. An evaluation on two datasets is conducted in \cref{section:evaluation}, while \cref{section:related-work} puts our work into perspective. Finally, \cref{section:conclusion} concludes our results.
\section{Background}
\label{section:background}
\paragraph{Condition Monitoring \& Predictive Maintenance}
Condition monitoring (CM) describes the field of monitoring and analyzing machines to deduct their current condition.
The condition of a machine is a critical information to plan a manual intervention or perform a maintenance task ahead of time, before a machine failure occurs.
Among a variety of approaches, modelling the normal behavior of the machine by incorporating sensor measurements into AI models has gained momentum in recent works \cite{stetco2019machine,kudelina2021trends, BeckerSSK22}. 
Often consisting of multiple sensory features such as temperature, vibration, or sound, the sensor data streams can be analyzed using machine learning to detect unexpected behavior in the machine, signaling a need to intervene. 
Points that deviate too far from the learned normal behavior can be viewed as anomalies in the data, rendering CM a subclass of the anomaly detection problem.
As the goal of CM is to support and improve the manual labor of domain experts, supervised learning approaches requiring fully labeled data are oftentimes not feasible in real-world applications.
This absence of labeled data poses a major challenge for machine learning models used in unsupervised CM, as they are not able to directly learn by comparison of their results against the ground truth of the data, in contrast to traditional machine learning architectures. Therefore, the normal behavior of the data is modeled and deviations are considered as anomalies.

\paragraph{Time series analysis}
The analysis of time series has been studied in many fields like monitoring cluster, machines, stock market analysis, and others \cite{9659499,yadav2020optimizing}.
Hereby, the sensor readings are analyzed with different goals like the modeling or prediction of future development of the time series where the application of ML has been widely adapted. 
Within this environment, Long Short-Term Memory neurons (LSTM) are especially suited for time-dependent contexts as they model the behavior of the process generating a time series by holding to the most relevant information.

\paragraph{Batch-wise Condition Monitoring}
\label{background:baseline}
In a similar work, Ahmad et al. \cite{kevinpaper} have proposed a CM approach on vibration data in combination with temperature measurements. 
Similar to our work, they propose a LSTM-Autoencoder based approach which exploits the time-dependence between batches to reconstruct the signal of high frequency accelerometer readings
and subsequently detect abnormal behavior. Their evaluation showed promising results and demonstrated the effectiveness of the proposed model. Nevertheless, they rely on transmitting all sensory 
data to a central server where the training is conducted on all monitored data, involving significant resources and therefore rendering it not feasible for an on-premise setup on lightweight edge devices in a remote industry site.

We consider this work as our baseline and aim to reduce the short-comings like a huge memory consumption that lies in the nature of considering multiple batches of sensor readings while training,
as well as high network utilization due to  transferring all training data to the cloud.

\section{Contribution}
\label{section:contribution}
\begin{figure*}
    \centering
    \includegraphics[width=\textwidth, keepaspectratio]{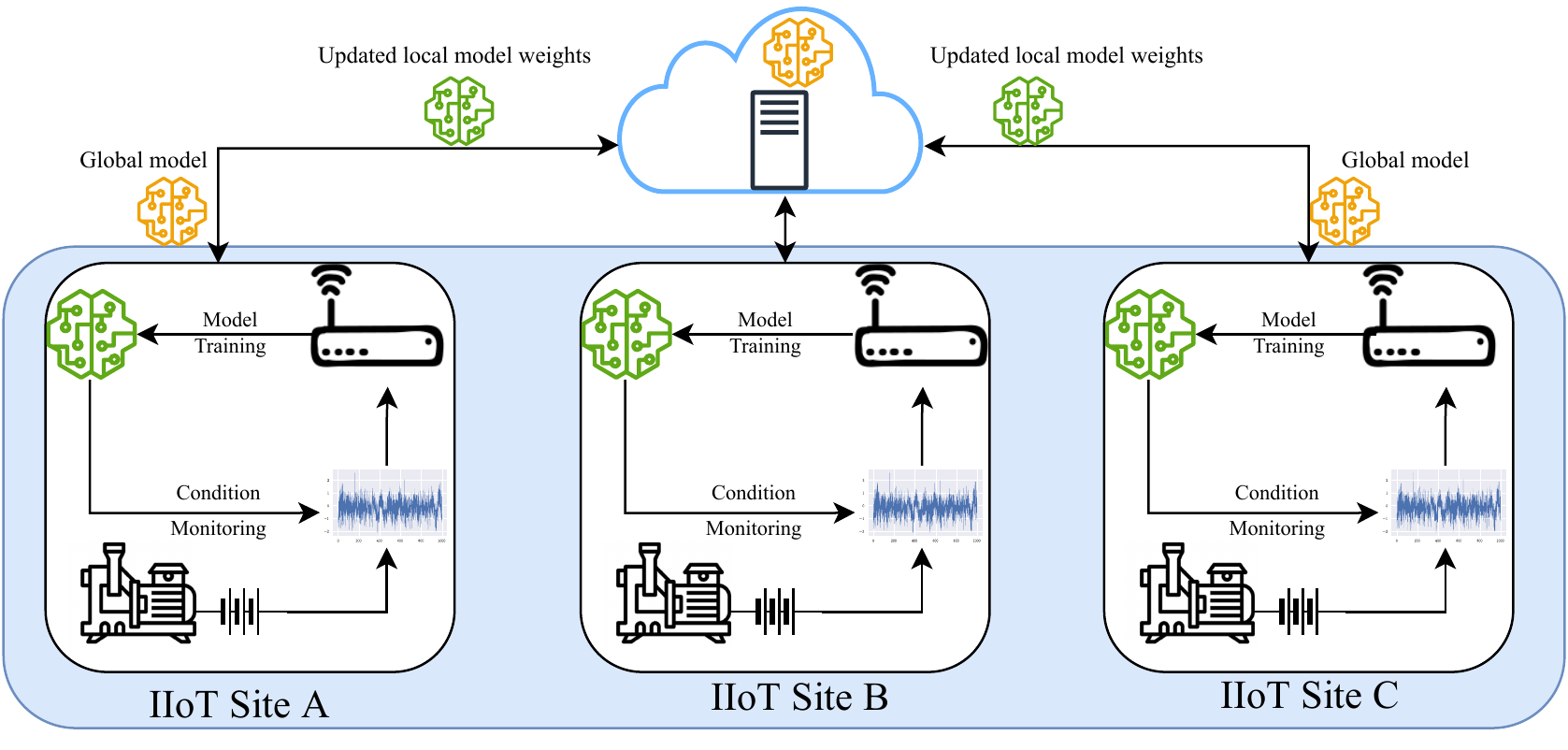}
    \caption{We envision an environment consisting of several IIoT sites, in which rotating machines are monitored with vibration sensors. Each site has one or multiple edge devices which train a model on the locally available sensor data and share the updated model weights with an aggregation node that in turn averages the weights to a global model that is finally distributed to the sites again.}
    \label{fig:environment}
\end{figure*}
We assume an environment as depicted in \cref{fig:environment}:
Multiple remotely located IIoT sites employ machinery with rotating components such as bearings, pumps or blades and are connected to the internet using a low-bandwidth connection.
Due to the wide availability of affordable sensors, these IIoT systems can be equipped with accelerometers, measuring i.e. vibrations of the machines.
Such measurements are often taken in batches, as a permanent recording is too costly in terms of energy management and detoriation of the sensor.
In addition, changes in the behavior of rotating machines, e.g. originating in the wear of a bearing, are expected to happen within days or hours and not within seconds.
Finally, the recorded vibration data can be used to detect a variety of possible anomalies like bearing faults, misalignments, or cavitation.

Therefore, we propose a FL-based approach to train the vibration data recorded on edge devices located in the IIoT site, which comes with several advantages and synergies:
The recorded data can stay locally as the training is not performed in the cloud.
Thus, the communication overhead is reduced to only share the model parameters instead of all recorded data.

In addition, anomaly data is very sparse for each individual machine.
This means that of all possible anomalies, only a small fraction on each individual machine will be observed within its life span.
Transferring learned knowledge about either normal or abnormal behavior between machines is an inherent strength of the FL approach as the global model aggregates modelled information from a variety of machines, thus sharing the knowledge of a multitude of sensor measurements in different sites or environments.

Based on these assumptions we state the following requirements for our approach:

\paragraph{Efficiency}
The proposed approach shall consist of a lightweight solution which is able to run on small edge devices.
Moreover, already existing environments shall be easily integrable and extendable in a cost-effective way.

\paragraph{Network and privacy awareness}
In order to save network resources, the data should remain local to decrease the communication overhead of transmitting sensor data over LTE or other low-bandwidth connections.
Additionally, the sensor data might contain business-sensitive information and should therefore not leave the organizational boundaries.

\paragraph{Model Effectiveness}
The performance of the condition monitoring FL implementation need to be competitive with state of the art centralized methods.

\paragraph{Deployability}
Considering the heterogeneous nature of IIoT environments, the implementation should be deployable across a variety of edge devices, regardless of i.e. the CPU architecture.

Finally, for this work, we assume the measurements to be largely equally distributed and independent and neglect further optimization of the FL approach itself:
For a real-world application, the amount of available data, the rate of measurements taken as well as the distribution of available datasets need to be taken into account while aggregating the different local models.

\subsection{Architecture}
\label{subsection:architecture}
\begin{figure*}[ht]
    \includegraphics[]{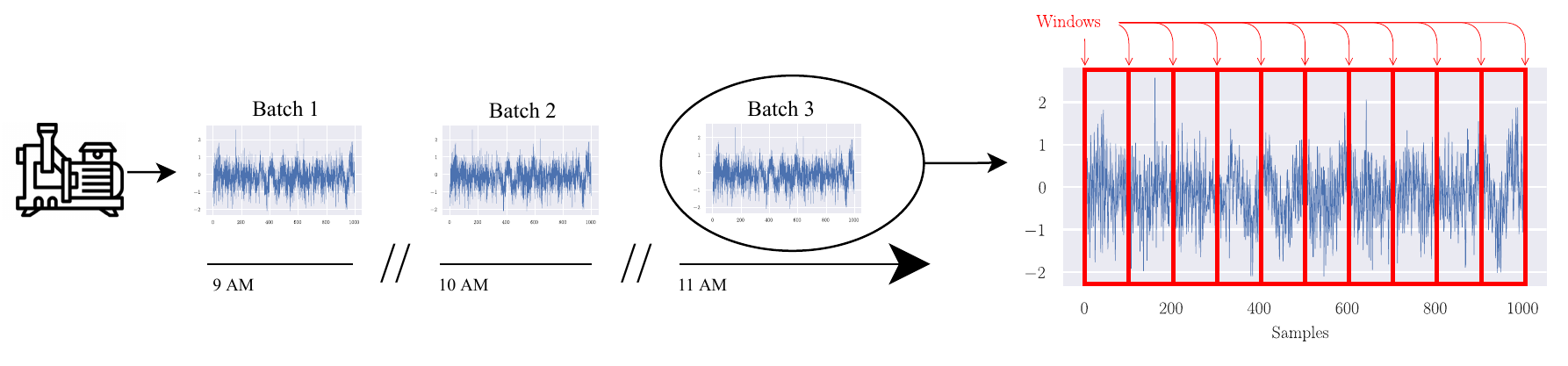}
    \caption{High level view on our approach: Batches, consisting of high-frequency vibration sensor measurements, are collected in specific time intervals, i.e. every hour, and then splitted into windows which are used as input for the model. Our methods works on single batches, instead of multiple preceding ones.}
    \label{fig:highlevel}
\end{figure*}
Each vibration measurement is a set of high-frequency accelerometer readings and defined by the length of the measurement and the sampling rate in Hertz (Hz).
A set of samples belonging to one measurement --  marked by a common timestamp --  is considered  a batch.
Therefore, as also depicted in \cref{fig:highlevel},  each batch is generated at a short interval and different batches are separated by larger time intervals like minutes or hours.
Following the previous work \cite{kevinpaper} described in \cref{section:background}, we adapt an LSTM Autoencoder, which aims to learn the normal behavior of all features in the data and has previously shown promising results.
The Autoencoder consists of two parts, an encoder and a decoder, which both consist of a variable number of sequentially aligned LSTM-Layers.
In our proposed Autoencoder, the input is time series data from the vibration measurements themselves.
The LSTM layers are used to provide the model with the ability to understand the sequential aspect of time series data \cite{siami2019performance}.
Thereby, the Autoencoder encodes a given measurement length of vibration data onto the size of the encoding from which the decoder learns to reconstruct healthy vibration data.

\begin{figure}
    \centering
    \includegraphics[width=\columnwidth, keepaspectratio]{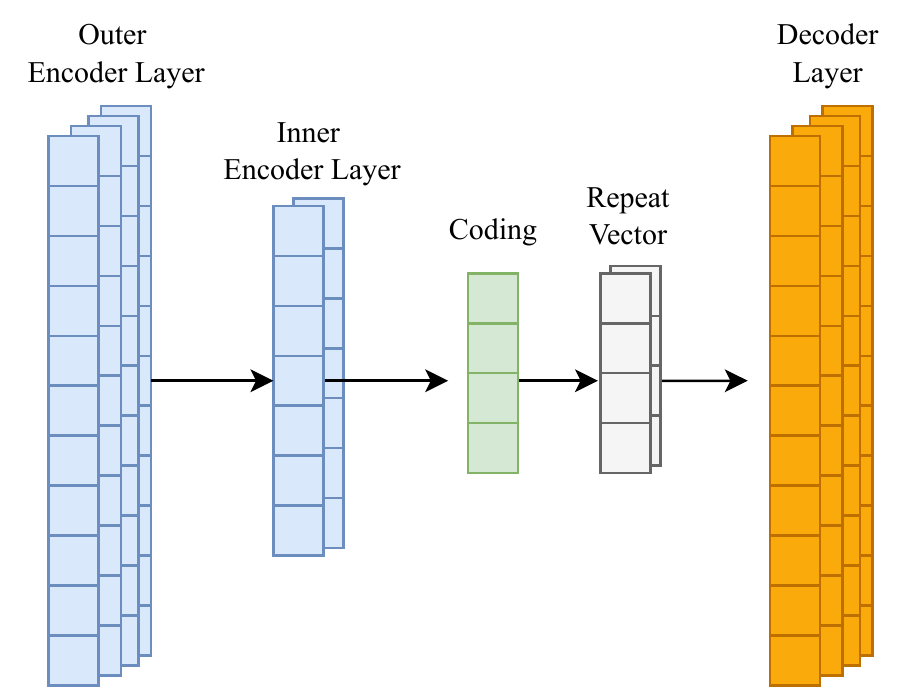}
    \caption{Architecture of the proposed model for the federated training nodes.}
    \label{fig:modelarchitecture}
\end{figure}
Contrary to the baseline method \cite{kevinpaper}, our proposed approach works only on the current measurement instead of multiple preceding batch measurements by exploiting the time-dependence of the data generating process within the vibrations of each batch. Consequently, as depicted in \cref{fig:highlevel}, we split the current batch into windows and test different window size configurations in \cref{subsection:hyperparameter}.
This change was made, as working on the level of single measurements requires significantly less memory usage compared to including multiple complete batch measurements into the CM model, and therefore directly benefits our \emph{efficiency} requirement.

The general architecture of the proposed method can be seen in \cref{fig:modelarchitecture}:
Internally, the LSTM layers pass their entire output sequence to the following layer to provide them with their full interpretation of the data.
In terms of the encoder, a final inner LSTM layer encodes the  output of the last LSTM-layer into the reduced encoding size.
For the decoder, the output of the encoding LSTM-layer is passed through the hidden LSTM layers into the final Dense layer which reconstructs the encoded time series.

To add non-linearity, Rectified Linear Units (ReLU) have been used and a L2 regularization has been applied with a value of $10^{-7}$ \cite{agarap2018deep}.
This regularisation penalizes the model for using excessively large weights or learning every aspect of the data, which is especially important in the case of small-scale vibration readings.

In order to detect anomalous and thus consequently deteriorating behavior, the prediction of the model is compared against the original input and used to compute a reconstruction error $RE_j$ for a given time period $j$, defined as the mean squared error (MSE) with original input signal $I$ and prediction $P$ with $I,P \in \mathbb{R}^{d}$ and the dimension of the measurement period $d$:
\begin{equation}
  RE_j = \frac{1}{d}\sum_{k=1}^{d} (I_k - P_k)^2
\label{eq:reconstruction_error}
\end{equation}

This RE, in the remainder of the paper also called \emph{anomaly score}, is used to evaluate the related time sequence in comparison to others by defining an accepted interval of values.
As the Autoencoder is trained on normal data, abnormal data is expected to deviate stronger.
To determine anomalies in the data, an \emph{anomaly threshold} defining this interval is introduced to differentiate between normal and abnormal behavior, with $\delta$ as a variable parameter indicating the sensitivity
as:

\begin{equation}
  \text{with }\overline{RE} = \frac{1}{d} \sum_{i=1}^{d} RE_{i}
\end{equation}

\begin{equation}
  RE \leq \delta * \sqrt{\frac{1}{d-1} \sum_{i=1}^d (RE_{i} - \overline{RE})^2}
\end{equation}

If the RE is smaller than this determined threshold, the current measurement is considered to have recorded normal data, while a larger value results in a detected anomaly.

\subsection{Hyperparameter Optimization}
\label{subsection:hyperparameter}
\begin{table}[h]
\centering
\caption{Model Hyperoptimization}
\begin{tabular}{lc}
    \toprule
    \multicolumn{2}{c}{\emph{Configuration and Search Space}}\\
    \midrule
    Optimizer & Adam\\
    Number Epochs & 100\\
    Batch size & \{32, 64, 128\}\\
    Window size & \{50, 100, 200\} \\
    Outer layer size & \{32, 64, 128, 256, 512\} \\
    Number Layers & \{1, 2, 3, 4 \} \\
    Hidden layer size & \{8, 16, 32\} \\
    Learning rate & \{$3*10^{-2}, 3*10^{-4}, 10^{-2}, 10^{-3}$\}\\
    \bottomrule
\end{tabular}
\label{tbl:hyperopt}
\end{table}

In regards to model optimization, we applied the Bayesian Optimization algorithm to test the performance of different hyperparameter combinations.
The search space of the Hyperparameter Optimization (HPO) is depicted in Table \ref{tbl:hyperopt}.
All models created using these parameter combinations were trained for 100 epochs.
The evaluation of the models is based on the validation loss of data excluded from the training and contains no anomalies.
Finally, a compromise of the smallest validation loss and smallest model size is made, depending on the dataset.

\subsection{Model Training}
As previously mentioned, the model is trained to reconstruct the normal vibration data accordingly.
Therefore, during the training the input and expected output of the model consists of the same time series chunks.
The goal of the optimization process for the Autoencoder is to be able to encode the essential information of this chunk into the hidden encoding layer, so that the original time series can subsequently be optimally reconstructed.
Thus, the Autoencoder is expected to learn the essential information constraining the normal behavior of the system under observation.

For the process of training our model, we use the Adam optimizer. %
In conjunction with the previously described kernel regularization of our LSTM layers, we use gradient clipping to prevent the exploding gradient problem \cite{zhang2019gradient}.
Further, we utilize an exponential learning rate scheduler which applies a stacking 1\% reduction to the learning rate at each epoch.
The learning rate itself is determined by the HPO described in \cref{subsection:hyperparameter} for the different datasets.

The loss function used for training and evaluation of the RE is the \textit{mean squared error} defined as
\begin{equation}
    MSE = \frac{1}{n} \sum_{j=1}^{n}RE_j
\end{equation}
with the RE as defined in Equation \ref{eq:reconstruction_error} and the total number of reconstructed time series windows $n$.
It penalizes prediction errors with an increasing scaling in severity.
This incentivizes the model to prioritize minimizing large prediction outliers, learning to predict the general oscillation of the vibration data, as small prediction errors are, due to the nature of the Autoencoder architecture, unavoidable.

\subsection{Federated Learning}
\label{subsection:federatedlearning}
\begin{figure}
    \centering
    \includegraphics[width=\columnwidth, keepaspectratio]{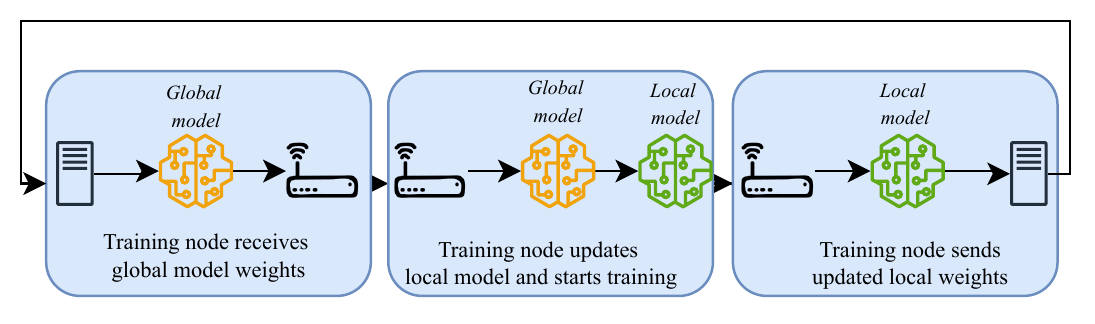}
    \caption{Lifecycle of a training node: The node receives the current global weights, uses them to initialize the local model and starts a training round for a given number of epochs on the local data. After the round, the updated local weights are send to the aggregator node.}
    \label{fig:training-cycle}
\end{figure}

\begin{figure}
    \centering
    \includegraphics[width=\columnwidth, keepaspectratio]{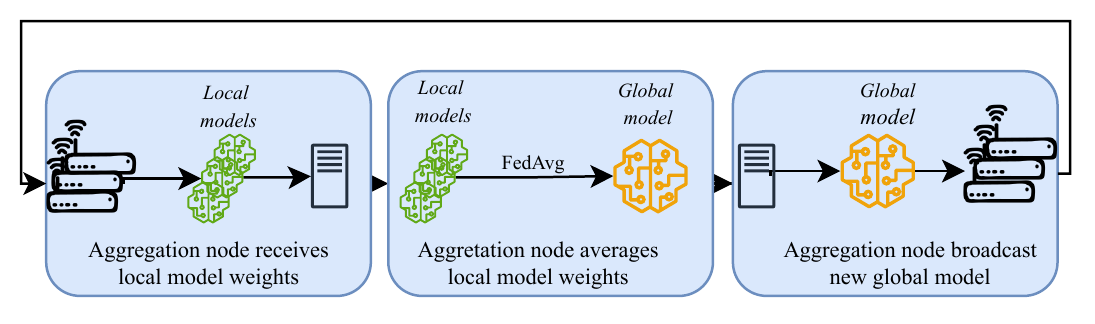}
    \caption{Lifecycle of the aggregation node: The node receives the updated model weights from the training nodes and averages them using the FedAvg algorithm. Finally, the new aggregated global weights are distributed to the training nodes again.}
    \label{fig:aggregation-cycle}
\end{figure}

In order to adapt the developed model to FL, in which devices located on-premise and close to the monitored machinery are utilized 
to collaboratively train a global model, we divide the environment in lightweight \textit{training nodes} and an \textit{aggregation node}, that are described in more detail in the remainder of this section and can be followed in \cref{fig:training-cycle,fig:aggregation-cycle}.

\subsubsection{Training Node}
The training node is a lightweight edge device in which the sensor data monitored from one or more machines arrives, i.e. via a message queue or data stream, and is consequently used to train a local model.
Following the first start, the device registers with the aggregation node, obtains the current global model, subsequently leverages it to further train 
on the locally observed data and finally applies it on the newly monitored sensor metrics to detect anomalies or deteriorating behavior of the respective monitored machinery.

After a configurable amount of training epochs, the training node shares the current local model weights with the aggregation node, where they are aggregated with the weights of other training workers.
Here, only the difference of the weights to the previously distributed global model is shared to preserve privacy during the communication.
The aggregated weights then constitute the new global model.
Finally, after the aggregation node has distributed the new global model weights to all participating training workers, the local models are re-instantiated with the respective weights and used for the condition monitoring process until a new model training is triggered.

As FL uses multiple training nodes, numerous instances of training workers exist and work in parallel, each training an instance of the same model architecture.

\subsubsection{Aggregation Node}
The aggregation node acts as the controlling instance of the FL approach.
It manages multiple training worker instances as clients, where newly connected worker instances are provided with the current state of the global model.
After completion of each training round, the global aggregation server receives model weight differences from every training node and aggregates them into the next iteration of the global model.
The aggregation of the received model weights is done via the FedAvg algorithm, as proposed by McMahan et al. \cite{mcmahan2017communication} where the received weight differences are first averaged and then applied to the global model.
It is chosen as recent research has shown the FedAvg algorithm to show good results compared to most state-of-the-art federated aggregation algorithms \cite{nilsson2018performance}.

With this approach, every training worker instance begins each iteration with an identical global model. 
After the next training round, the local model instances only differ in their model weights by the difference introduced through their training data during this iteration.
This continuous re-integration and aggregation of the local model instances allows for the general training to jointly converge towards the shared global optimum observed over the entirety of the training data available on the different IIoT nodes.

\section{Evaluation}
\label{section:evaluation}
We conducted multiple experiments to evaluate the FL approach in general and compared it against the baseline model presented in \cref{section:background}.
To simulate a real-life use case scenario for our evaluation, we leverage two real-world datasets consisting of vibration sensor data monitored from multiple machines. The datasets are described in more detail in the next section. 

In addition, the evaluation was conducted in three different testbed settings: A \emph{central server}, a \emph{virtualized testbed} and an \emph{edge testbed} respectively, consisting of the devices listed in \cref{tab:devices}. The models were provided in multi-arch Docker containers in order to enable a deployment across heterogeneous devices and consequently fulfill the \emph{deployability} requirement stated in \cref{section:contribution}.

As evaluation metrics we utilize the effectiveness of the models, the respective resource utilization, as well as the general runtime across
different devices and settings.

\begin{table}[]

\caption{Devices used for the evaluation.}
\label{tab:devices}
\resizebox{\columnwidth}{!}{%
\begin{tabular}{@{}cccc@{}}
\toprule
              & Central server                                                  & Virtualized Testbed                                               & Edge Testbed                                                     \\ \midrule
Machine types & Commodity server                                                & \begin{tabular}[c]{@{}c@{}}GCP VM \\ (e2-medium)\end{tabular}     & Raspberry Pi 4B                                                  \\
CPU           & \begin{tabular}[c]{@{}c@{}}AMD EPYC 7282\\  2.8GHz\end{tabular} & \begin{tabular}[c]{@{}c@{}}Intel Xeon CPU \\ 2.2 GHz\end{tabular} & \begin{tabular}[c]{@{}c@{}}ARM Cortex-A72\\  1.5GHz\end{tabular} \\
Cores         & 32                                                              & 1 vCPU                                                            & 4vCPU                                                            \\
Memory        & 128 GB                                                          & 4 GB                                                              & 4 GB                                                             \\
                                                        \bottomrule
\end{tabular}%
}
\end{table}

\subsection{Dataset descriptions}
\label{subsection:dataset-description}
For the evaluation we were using two different real-world datasets, explained in more detail in this section.
\paragraph{Bearing Dataset}
The first dataset is provided by \emph{NASA IMS}, publicly available, and contains monitored vibration sensor metrics of four bearings \cite{lee2007rexnord}.
The data was measured for 1 second every 10 minutes with a sampling rate of 20,480 Hertz, and the whole dataset consists of three independent run-to-failure experiments (\emph{Set1-3}) in which the bearings were monitored until a failure occurred.
Therefore, at the end of each experiment, a subset of the monitored bearings develops a degrading health state.
\cref{tab:nasadata} shows the properties of the collected data and the faulty bearings for every experiment.
Because of the very high-precision sampling rate, a downsampling resulting in a sample rate of 4096Hz has been applied.
\begin{table}[]
\centering
\caption{Overview of different experiments of IMS bearing vibration dataset.}
\label{tab:nasadata}
\begin{tabular}{@{}cccccc@{}}
\toprule
Experiment & \begin{tabular}[c]{@{}c@{}}Number\\  Batches\end{tabular} & \begin{tabular}[c]{@{}c@{}}Sampling \\ Rate {[}Hz{]}\end{tabular} & \begin{tabular}[c]{@{}c@{}}Batch \\ Size \end{tabular}  & \begin{tabular}[c]{@{}c@{}}Faulty\\  Bearings\end{tabular} & \begin{tabular}[c]{@{}c@{}}Resampled\\  Size\end{tabular}  \\ \midrule
Set1         & 2156                                                           & 20480                                & 20480                             & B3 \& B4   & 4096                                                \\
Set2         & 984                                                            & 20480                                & 20480                             & B1         & 4096                                                \\
Set3         & 6324                                                           & 20480                                & 20480                             & B3         & 4096                                                \\ \bottomrule
\end{tabular}
\end{table}

\paragraph{Rotating Machine Dataset}
This dataset was provided by an industry partner and contains vibration metrics collected from five different rotating machines (RM).
The vibration data was measured for three dimensions (x,y,z) over time, by utilizing an accelerometer sensor attached to the housing of the machine \cite{kevinpaper}.
As in the previous dataset, the data was collected in batches and measured hourly. Additionally, the data was labeled depending on the condition of the machine during that time with either anomalous or normal behavior, albeit the labels were only used for the evaluation.

\begin{table}[]
\centering
\caption{Overview of different experiments of industry vibration dataset.}
\begin{tabular}{ccccc}
\toprule
Machine & \begin{tabular}[c]{@{}c@{}}Number\\ Batches\end{tabular} & \begin{tabular}[c]{@{}c@{}}Sampling\\ Rate [Hz]\end{tabular} & Batch Size & Anomalies \\ \midrule
RM-1    & 1176                                                     & 4096                                                    & 10240      &    53       \\
RM-2    & 1463                                                     & 4000                                                    & 800        &     4      \\
RM-3    & 2204                                                     & 4000                                                    & 800        &     4      \\
RM-4    & 1452                                                     & 4000                                                    & 800        &     4      \\
RM-5    & 1452                                                     & 4000                                                    & 800        &     4      \\ \bottomrule
\end{tabular}
\label{tab:industry-dataset}
\end{table}

Although the machines are build-wise unique, they are from the same type which allows for a knowledge transfer between them in a federated learning setting. \cref{tab:industry-dataset} shows the characteristics of the dataset. The data is measured with a sampling rate of 4000Hz for 0.2 seconds, resulting in 800 data points per batch with a varying amount of abnormal batches labeled for the different rotating machines.

\subsection{Experiment setup}
For the following experiments we distributed the datasets described in the previous section to the nodes mentioned in \cref{tab:devices}.
In the \emph{central server} setting we start the aggregator and training worker in different containers but on the same node, whereas we launch four respectively five different virtual machines in the Google Cloud Platform as training workers for the different bearings and rotating machines.

As parameter for the model training we use the following, obtained from the HPO presented in \cref{subsection:hyperparameter}: A  \emph{window size} of 100, \emph{learning rate} of 0.001, as well as one outer LSTM-layer with a size of 128, a \emph{hidden layer size} of 16, and a \emph{batch size} of 64.
Although we found that an LSTM layer size of 512 would have been optimal, we decided to use 128 layers as it achieved almost equivalent results while being significantly more efficient in terms of computation.

We conducted experiments with two different data availability scenarios: First, we assume a scenario in which historical sensor data is available. Therefore, during model training
the training worker use the whole dataset, split into 70\% training data from which 8\% are used as validation data and the rest for testing.
Furthermore, also a \emph{cold-start} scenario is considered, in which we do not assume any historical data. Therefore, for each federated training round, the amount of available training windows increases, simulating newly arriving measurement data.

For all experiments -- except the \emph{cold-start} scenario -- we applied 25 federated training rounds in which each training node trained for a single epoch before aggregating the weights, and compared it against the baseline model which was trained on all available training data for 100 epochs.

\subsection{General applicability}
\begin{figure}
    \centering
    \includegraphics{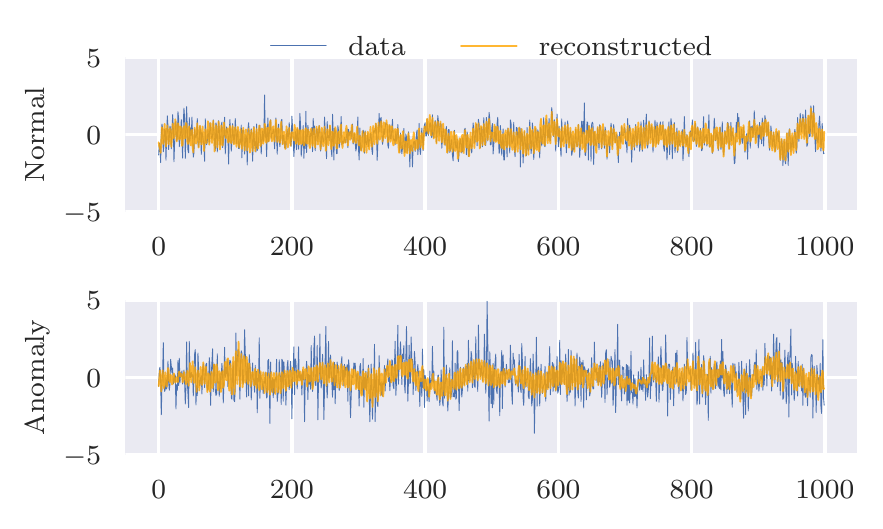}
    \caption{Original data and reconstruction of a healthy and a defective bearing in \emph{Set2} of the IMS bearing dataset. For the anomalous bearing the reconstructed signal differs significantly from the original data, hence indicating an anomaly.}
    \label{fig:e2chunk}
\end{figure}
To assess the general applicability for anomaly detection of the proposed approach, we compare the reconstructed signal of a healthy and faulty measurement.

Figure \ref{fig:e2chunk} visualizes two measurement periods of the same bearing in the IMS bearing dataset, during normal behavior and subsequently during an anomalous phase:
In blue we depict the original vibration sensor readings and the orange part shows the reconstructed signal by our model.
As can be seen, both vibration periods are of different amplitudes, with the anomalous data having a maximum amplitude of about two times the normal one.
While our approach is able to reconstruct the normal vibration data rather closely, the reconstruction error increases significantly during an anomaly, hence allowing for a classification based on our model.
The model has shown that the structure of normal data is learned while the reconstruction of anomalous data is performing worse.

\subsection{Comparison between baseline and federated model}

In this section, we compare our achieved results to the baseline by Ahmad et al. \cite{kevinpaper} described in \cref{section:background}.
First, we investigate the performance of our approach on the first dataset in terms of detection performance and subsequently, we compare the accuracy of the models for the industry dataset.

\begin{figure}[t]
    \centering
    \includegraphics{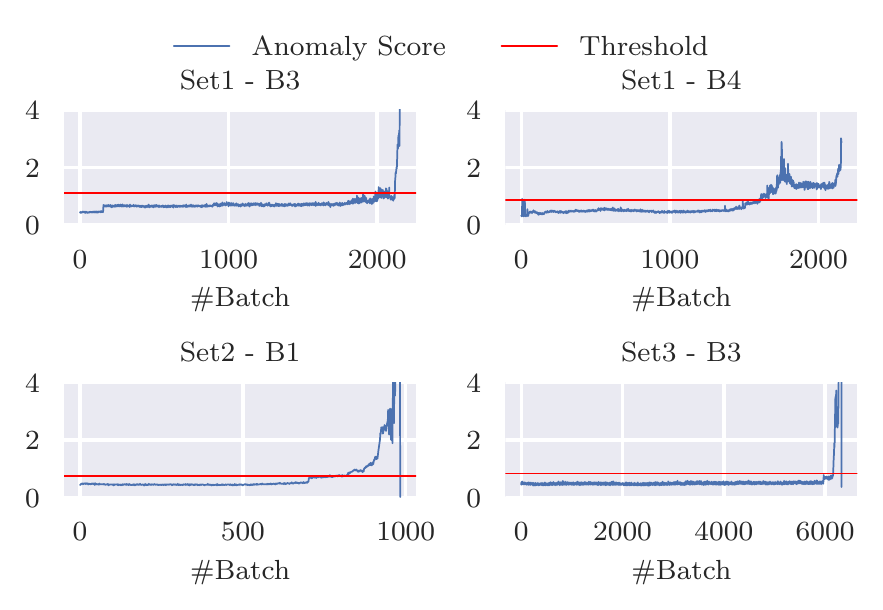}
    \caption{Anomaly predictions by the Federated Learning model for all different bearing faults in the \emph{IMS bearing dataset}. A deteriorating behavior was detected as soon as the anomaly scores exceeded the threshold.}
    \label{fig:fedfaultynasa}
\end{figure}
\begin{figure}
    \centering
    \includegraphics{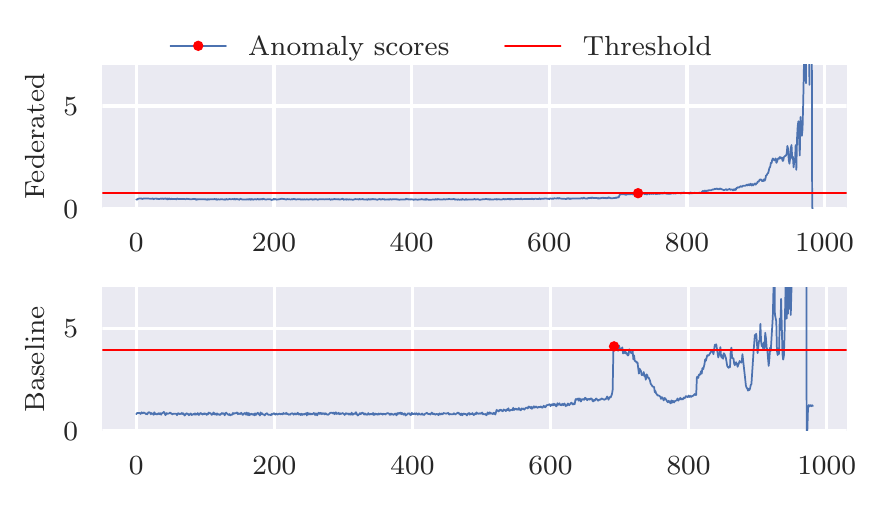}
    \caption{Comparison between anomaly scores and threshold of the federated and baseline models for a faulty bearing. Both models correctly detect deteriorating behavior at the end of the experiment.}
    \label{fig:E2-0_RE}
\end{figure}

\begin{figure}
    \centering
    \includegraphics{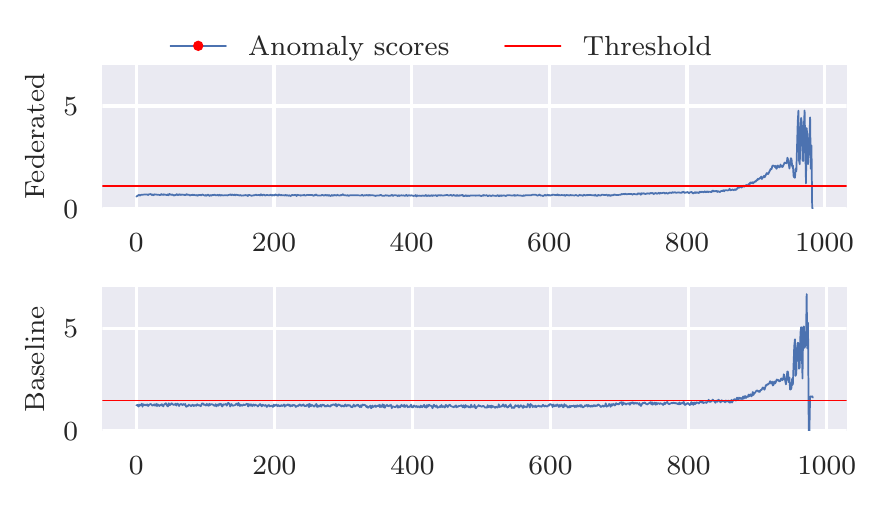}
    \caption{Comparison between anomaly scores and threshold of the federated and baseline models for a healthy bearing. No false positives were detected at the beginning of the experiment and only the propagating anomaly from another bearing was detected at the end of the experiment.}
    \label{fig:E2-1_RE}
\end{figure}
\begin{figure*}
    \centering
    \includegraphics{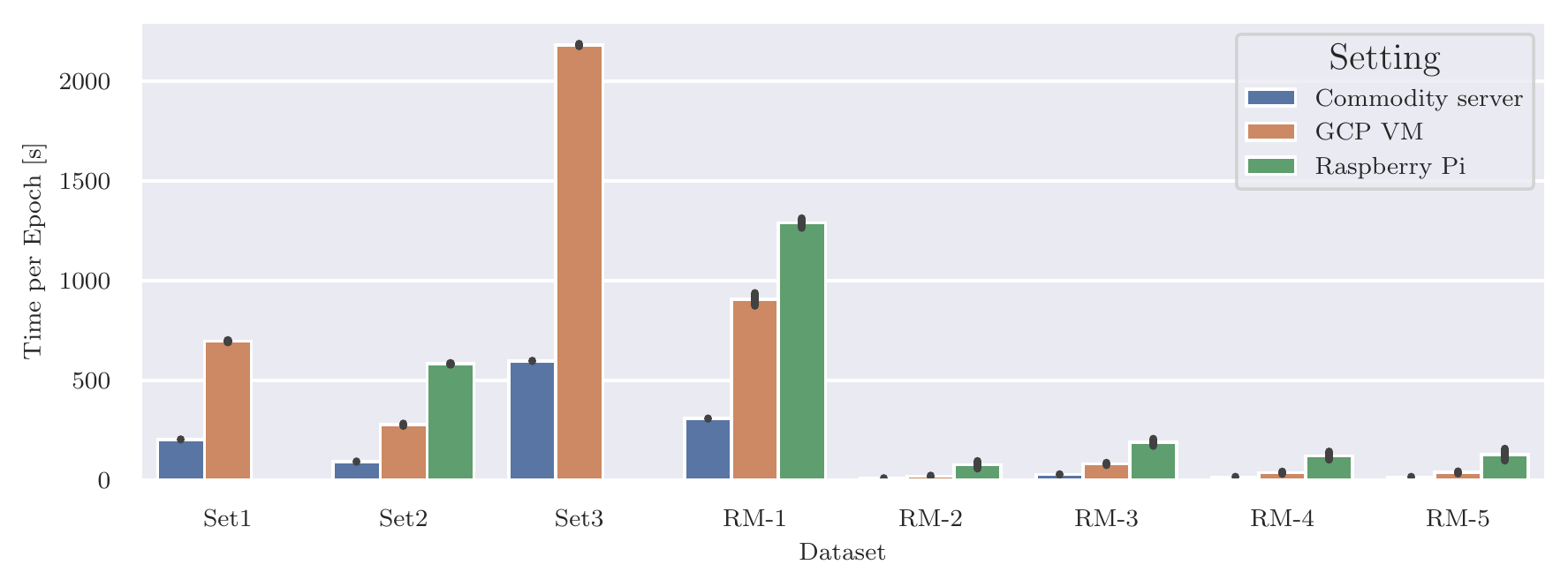}
    \caption{Average training times per epoch for all used datasets and evaluation settings when training on the whole datasets.}
    \label{fig:epoch_times}
\end{figure*}
For the IMS dataset, the model was trained on the initial healthy data of each bearing, assuming that historical data is available.
In Figure \ref{fig:fedfaultynasa}, we can see the reconstruction error values and, therefore, predictions of the condition of the bearing for all faulty bearings of the experiment compared.
In every case, the federated learning model predicted a defect in the final stages of the experiment, with a strong increase in the RE.
Simultaneously, the early stages of the experiment were predicted as normal up until the time of the detected anomaly.
No clear false positives were predicted in the early stages of the experiments, for which we know that no defects were present.

Afterward, the detection of the faulty bearing \emph{Set2} is compared to the baseline method.
In \cref{fig:E2-0_RE} the anomaly scores of the first bearing of \emph{Set2} are depicted in dependence of the measured batch number during the duration of the experiment.
Our approach detects a deteriorating behavior at a comparable point in time to the baseline model.
In addition, for both models the anomaly scores rise significantly towards the end of the experiment, indicating a similar performance.

Moreover, \cref{fig:E2-1_RE} depicts the anomaly score of the second bearing of \emph{Set2}:
This bearing had no failure during the experiment and showcases how the anomaly detector behaves for a healthy bearing.
It can be observed that the model reliably accepts normal data without indicating anomalies.
Only towards the end of the experiment, the vibration of the strongly anomalous behaving first bearing propagates to the second bearing, explaining the increase of anomaly scores.

For the industry dataset, the anomaly detection performance of the proposed approach in comparison to the baseline can be seen in \cref{tab:comparison-federated-baseline}. 
The results for F1 score, precision, and recall are nearly identical for the majority.
As the proposed approach works on smaller windows of the associated time series, our FL trained model is able to predict for a larger group of rotating machines.
Thus, we could include the RM-1 although it initially uses a different batch size, since the proposed model is not limited by it, contrary to the baseline.
Evidently, it is important to mention that only very few anomalies exists in the RM datasets, resulting in an average F1-score of 99.4\%. 
Nevertheless, the achieved results are encouraging as the performance of the model is very competitive despite the much lower complexity compared to the baseline model.

\subsection{Knowledge Transfer}
To test the knowledge transfer capabilities of our approach, we used the resulting global model of \emph{Set2} after a federated model training with 25 epochs and evaluated it against a faulty bearing in the first experiment set (\emph{Set1}), therefore simulating a warm-start in another IIoT faculty for a bearing of the same type.
\begin{figure}
    \includegraphics{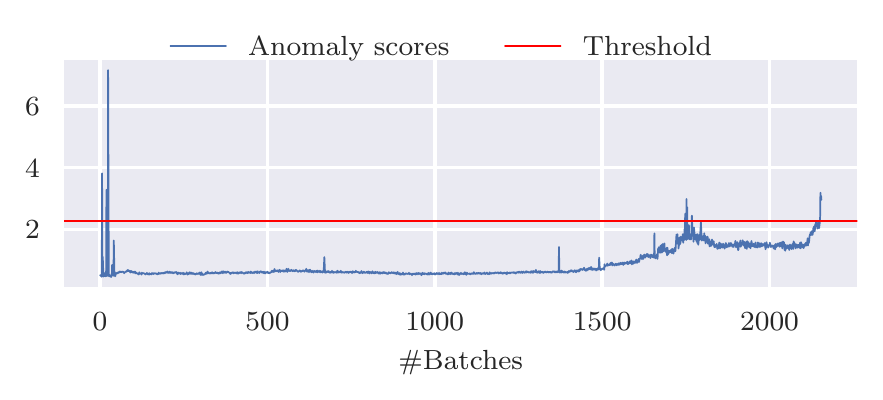}
    \caption{Anomaly scores and threshold results from Bearing 4 in \emph{Set1}, from a model which was was only trained on \emph{Set2}, representing knowledge transfer between participating industry sites.}
    \label{fig:transfer-learning}
\end{figure}

\cref{fig:transfer-learning} shows the reconstruction errors and threshold of this experiment. As can be seen, the model of \emph{Set2} is already able to detect the deteriorating behavior at the end of the experiment, although it was only trained on data from another experiment with significantly less available data points.

\subsection{Runtime and resource utilization}
For the IIoT use case, the resource utilization is a major limiting factor.
Different hardware configurations have been tested for their needed time per training epoch, as depicted in \cref{fig:epoch_times}.
As expected the training times for the commodity server are the lowest as it comes with the strongest CPU, followed by the GCP virtual machines, and the Raspberry Pis that are representative of edge nodes. 
Unfortunately, the whole dataset of the Set1 and Set3 dataset of the bearing dataset did not fit into the memory of the Raspberry Pi, therefore rendering the training on the whole dataset unfeasible.

\begin{figure}[t]
    \includegraphics{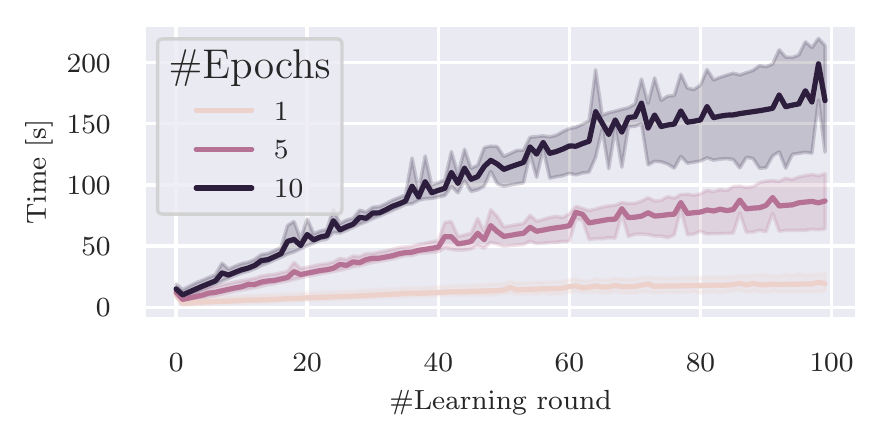}
    \caption{Average training time per round for RM 1-5 and different amounts of training epochs per round. The plot depicts the cold-start setting where the amount of available training windows increases per round, i.e. in round 10 the model uses 640 windows to train on and 6400 windows in round 100, respectively.}
    \label{fig:fed_round_times}
\end{figure}

Therefore, the \emph{cold-start} scenario has been implemented.
Here, for every following epoch the number of available training windows is increased, reproducing a real-world scenario where more and more data is measured.
The first epoch starts with 64 windows, according to the $batch size$ of the LSTM Autoencoder training, for every subsequent epoch this is increased by further 64 windows.
The results can be seen in \cref{fig:fed_round_times} for different configurations of how many epochs were trained per round.
The performance measures thereby achieved similar results to the full training runs.
Thus, also Raspberry Pis are able to train with the amount of data in a reasonable amount of time.
For example, the NASA bearing dataset delivered measurements every 10 minutes, whereas the computation time even for learning round 100 with 6400 windows lies well within this interval, rendering an online measuring and learning scenario feasible with the proposed approach.

\begin{figure}
    \includegraphics{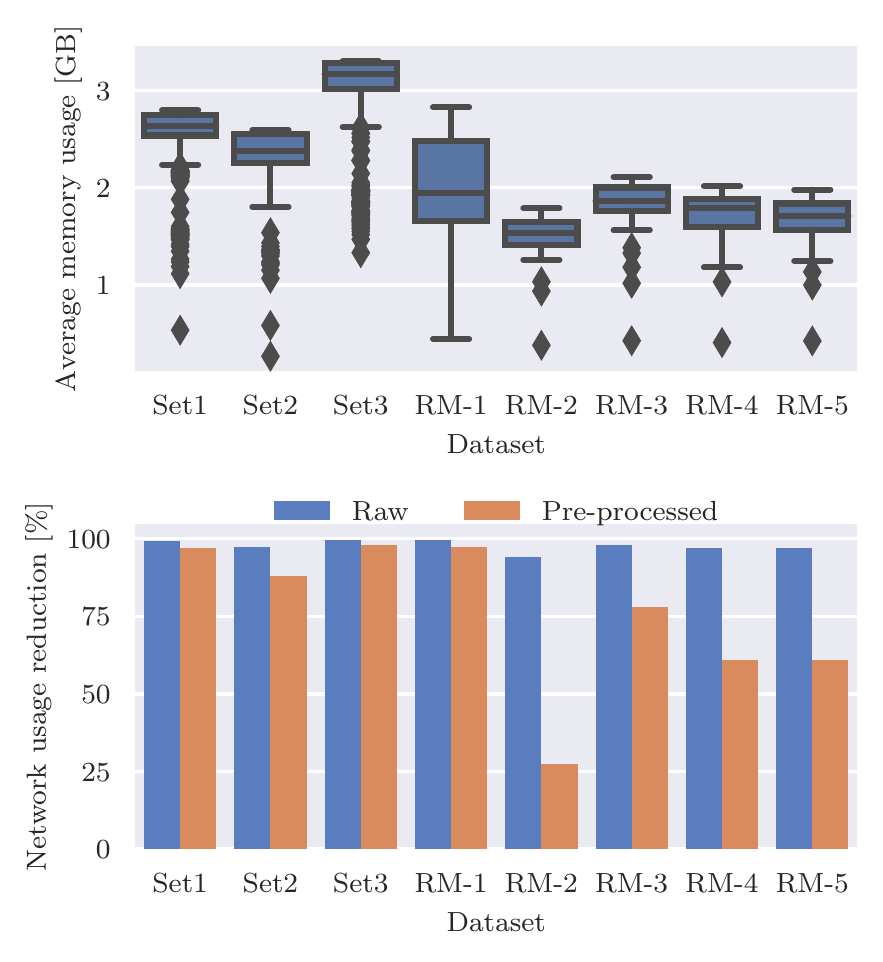}
    \caption{Results of the resource utilization evaluation. The memory usage was monitored on the respective training nodes while training on the datasets and for \emph{Set1-3} the average usage over all four bearings is shown. The network usage reduction was calculated by utilizing the dataset sizes before and after pre-processing (only for the actual vibration readings) and comparing it to the weight transmissions in case of the federated model. 
}
    \label{fig:resource-usages}
\end{figure}

In addition, the resource consumption of the proposed approach is compared to the baseline, as can be seen in \cref{fig:resource-usages}.
Depending on the size of the dataset, the FL workers require between 2 and 3GB of main memory to conduct a local model training. In comparison, 
the baseline needs about 9 to 11GB of memory in average per dataset, thus rendering it not feasible for the execution on IIoT devices.
Even more significant is the reduction in overall network utilization: In case of the FL approach, only model weights have to be communicated between the training and aggregation nodes, resulting in around 6.3MB for each tested machine and 25 training rounds.
The baseline requires transmission of all training data to the central server, depicted as \emph{Raw} in \cref{fig:resource-usages}, which in consequence enables a reduction of the overall network usage from up to 99.2\%. We further also show a scenario where the data is pre-processed in the IIoT site (i.e. resampled and additional information such as temperature measurements are removed) and the FL approach is still able to significantly outperform the baseline or other centralized methods.

Summarizing, the results show that the proposed method is able to achieve the requirements stated in \cref{section:contribution}.

\addtolength{\tabcolsep}{-3pt}
\begin{table}[]
\centering
\caption{Model performance comparison between the federated and baseline models on the industry dataset.}
\begin{tabular}{|lclcclcclc|}
\hline
                              & \multicolumn{3}{c}{F1 Score}                                  & \multicolumn{3}{c}{Precision}                                 & \multicolumn{3}{c|}{Recall}                                   \\ \hline
\multicolumn{1}{|l|}{Dataset} & \multicolumn{2}{l}{Federated} & \multicolumn{1}{l|}{Baseline} & \multicolumn{2}{l}{Federated} & \multicolumn{1}{l|}{Baseline} & \multicolumn{2}{l}{Federated} & \multicolumn{1}{l|}{Baseline} \\ \hline
\multicolumn{1}{|l|}{RM-1}    & \multicolumn{2}{c}{0.970}     & \multicolumn{1}{c|}{-}        & \multicolumn{2}{c}{0.942}     & \multicolumn{1}{c|}{-}        & \multicolumn{2}{c}{1.000}     & -                             \\
\multicolumn{1}{|l|}{RM-2}    & \multicolumn{2}{c}{1.000}     & \multicolumn{1}{c|}{1.000}    & \multicolumn{2}{c}{1.000}     & \multicolumn{1}{c|}{1.000}    & \multicolumn{2}{c}{1.000}     & 1.000                         \\
\multicolumn{1}{|l|}{RM-3}    & \multicolumn{2}{c}{1.000}     & \multicolumn{1}{c|}{0.997}    & \multicolumn{2}{c}{1.000}     & \multicolumn{1}{c|}{0.997}    & \multicolumn{2}{c}{1.000}     & 0.997                         \\
\multicolumn{1}{|l|}{RM-4}    & \multicolumn{2}{c}{1.000}     & \multicolumn{1}{c|}{1.000}    & \multicolumn{2}{c}{1.000}     & \multicolumn{1}{c|}{1.000}    & \multicolumn{2}{c}{1.000}     & 1.000                         \\
\multicolumn{1}{|l|}{RM-5}    & \multicolumn{2}{c}{1.000}     & \multicolumn{1}{c|}{1.000}    & \multicolumn{2}{c}{1.000}     & \multicolumn{1}{c|}{1.000}    & \multicolumn{2}{c}{1.000}     & 1.000                         \\ \hline
\end{tabular}
\label{tab:comparison-federated-baseline}
\end{table}
\addtolength{\tabcolsep}{3pt}

\section{Related Work}
\label{section:related-work}
In terms of general condition monitoring a variety of related work exist. For instance,
Sakib et al. \cite{sakib2020migrating} achieved low-latency condition monitoring inference by deploying pre-trained models to IIoT devices. In their approach, the models are trained on centralized servers and the IIoT devices are only used for inference.
Hahn and Mechefske \cite{hahn2021self} utilized a self-supervised approach of Condition Monitoring using a Variational-Autoencoder on the NASA milling data set, which does not require feature engineering and shows promising results on some parameters.
In addition, Mostafavi and Sadighi \cite{mostafavi2021novel} proposed a condition monitoring framework which is fully located on IIoT devices.
Their framework, like ours, uses an LSTM-Autoencoder for anomaly detection while being resource-restricted.
These approaches, while overcoming the resource restriction, lack the advantages that FL has to offer, as they do not keep the data in the local sites nor do they share information between multiple devices.

In regards to FL approaches, Wang et al. and Yunfan et al. developed two generalized frameworks, In-Edge AI \cite{wang2019edge} and EdgeFed\cite{ye2020edgefed}, for improving FL at the edge. They explore different approaches to reduce the system communication load and partly offload resource-intensive work to the cloud to archive a more effective aggregation. 
Wu et al. \cite{wu2020personalized} proposed a framework for mitigating issues of strong heterogeneity by the usage of personalized FL methods.
Dhada et al. \cite{dhada2020federated} used a FL approach for determining the remaining useful life of engines on a simulated turbofan data set, a topic strongly related to the basic idea of condition monitoring also using rotating machines.
Our work, in contrast, deals with the application of such a use case while considering the resource limitation of edge devices, especially within an IIoT environment.
Zhang et al. \cite{zhang2020blockchain} developed a blockchain-based FL solution for condition monitoring in the IIoT.
Their approach included a novel weighted federated averaging algorithm to mitigate the data heterogeneity issue and the utilization of smart contracts for client incentives.
We neglect utilizing a blockchain approach, since it would introduces new additional computational overhead.

\section{Conclusion}
\label{section:conclusion}
In this paper, we presented an Autoencoder-based Federated Learning approach for efficient condition monitoring of rotating machines in IIoT environments, utilizing monitored vibration data.
Our method enables a collaborative and distributed model training on lightweight edge devices that are located in close proximity to the sensor data sources in industrial sites.
While first learning to reconstruct the normal behaviour of the monitored sensor streams, it allows for an unsupervised anomaly detection based on reconstruction errors.

The conducted evaluation on two real-world datasets shows a competitive performance compared to a state of the art baseline and at the same time decreases the complexity enough to qualify low-powered edge devices as training nodes. Moreover, the network utilization is decreased by up to 99.22\%, further benefiting remote industry sites with possibly unreliable or expensive network connections.

In future work, we plan to evaluate our method on additional datasets and to test further weight averaging algorithms.

\section*{Acknowledgement}
\footnotesize
This work is supported through the Deutsche Forschungsgemeinschaft (DFG, German Research Foundation) as FONDA (Project 414984028, SFB 1404) and HEIBRIDS - Helmholtz Einstein International Berlin Research School in Data Science under contract no. HIDSS-0001.

\bibliographystyle{IEEEtran}
\bibliography{iiot}

\begin{thebibliography}{10}
\providecommand{\url}[1]{#1}
\csname url@samestyle\endcsname
\providecommand{\newblock}{\relax}
\providecommand{\bibinfo}[2]{#2}
\providecommand{\BIBentrySTDinterwordspacing}{\spaceskip=0pt\relax}
\providecommand{\BIBentryALTinterwordstretchfactor}{4}
\providecommand{\BIBentryALTinterwordspacing}{\spaceskip=\fontdimen2\font plus
\BIBentryALTinterwordstretchfactor\fontdimen3\font minus
  \fontdimen4\font\relax}
\providecommand{\BIBforeignlanguage}[2]{{%
\expandafter\ifx\csname l@#1\endcsname\relax
\typeout{** WARNING: IEEEtran.bst: No hyphenation pattern has been}%
\typeout{** loaded for the language `#1'. Using the pattern for}%
\typeout{** the default language instead.}%
\else
\language=\csname l@#1\endcsname
\fi
#2}}
\providecommand{\BIBdecl}{\relax}
\BIBdecl

\bibitem{FerrerB0TK21}
A.~J. Ferrer, S.~Becker, F.~Schmidt, L.~Thamsen, and O.~Kao, ``Towards a
  cognitive compute continuum: An architecture for ad-hoc self-managed
  swarms,'' in \emph{CCGRID}.\hskip 1em plus 0.5em minus 0.4em\relax {IEEE},
  2021.

\bibitem{Sari2020}
A.~Sari, A.~Lekidis, and I.~Butun, \emph{Industrial Networks and IIoT: Now and
  Future Trends}.\hskip 1em plus 0.5em minus 0.4em\relax Springer, 2020.

\bibitem{kevinpaper}
S.~Ahmad, K.~Styp-Rekowski, S.~Nedelkoski, and O.~Kao, ``Autoencoder-based
  condition monitoring and anomaly detection method for rotating machines,'' in
  \emph{Big Data}.\hskip 1em plus 0.5em minus 0.4em\relax IEEE, 2020.

\bibitem{carden_vibration_2004}
E.~P. Carden and P.~Fanning, ``Vibration based condition monitoring: A
  review,'' vol.~3, no.~4, 2004, {SAGE} Publications.

\bibitem{becker_towards_2020}
S.~Becker, F.~Schmidt, A.~Gulenko, A.~Acker, and O.~Kao, ``Towards {AIOps} in
  edge computing environments,'' in \emph{Big Data}.\hskip 1em plus 0.5em minus
  0.4em\relax IEEE, 2020.

\bibitem{wang_toward_2019}
G.~Wang, M.~Nixon, and M.~Boudreaux, ``Toward cloud-assisted industrial {IoT}
  platform for large-scale continuous condition monitoring,'' vol. 107, no.~6,
  2019, proceedings of the {IEEE} 107.

\bibitem{m_fusion_2021}
P.~Boopalan, S.~P. Ramu, Q.-V. Pham, K.~Dev, P.~K.~R. Maddikunta, T.~R.
  Gadekallu, T.~Huynh-The \emph{et~al.}, ``Fusion of federated learning and
  industrial internet of things: A survey,'' \emph{Computer Networks}, p.
  109048, 2022.

\bibitem{fed-google}
J.~Kone{\v{c}}n{\`y}, H.~B. McMahan, F.~X. Yu, P.~Richt{\'a}rik, A.~T. Suresh,
  and D.~Bacon, ``Federated learning: Strategies for improving communication
  efficiency,'' \emph{arXiv preprint arXiv:1610.05492}, 2016.

\bibitem{stetco2019machine}
A.~Stetco, F.~Dinmohammadi, X.~Zhao, V.~Robu, D.~Flynn, M.~Barnes, J.~Keane,
  and G.~Nenadic, ``Machine learning methods for wind turbine condition
  monitoring: A review,'' \emph{Renewable energy}, vol. 133, 2019.

\bibitem{kudelina2021trends}
K.~Kudelina, T.~Vaimann, B.~Asad, A.~Rass{\~o}lkin, A.~Kallaste, and
  G.~Demidova, ``Trends and challenges in intelligent condition monitoring of
  electrical machines using machine learning,'' \emph{Applied Sciences},
  vol.~11, no.~6, 2021.

\bibitem{BeckerSSK22}
S.~Becker, D.~Scheinert, F.~Schmidt, and O.~Kao, ``Efficient runtime profiling
  for black-box machine learning services on sensor streams,'' in
  \emph{ICFEC}.\hskip 1em plus 0.5em minus 0.4em\relax {IEEE}, 2022.

\bibitem{9659499}
S.~Becker, F.~Schmidt, L.~Thamsen, A.~J. Ferrer, and O.~Kao, ``Los:
  Local-optimistic scheduling of periodic model training for anomaly detection
  on sensor data streams in meshed edge networks,'' in \emph{ACSOS}.\hskip 1em
  plus 0.5em minus 0.4em\relax IEEE, 2021.

\bibitem{yadav2020optimizing}
A.~Yadav, C.~Jha, and A.~Sharan, ``Optimizing lstm for time series prediction
  in indian stock market,'' \emph{Procedia Computer Science}, vol. 167, 2020.

\bibitem{siami2019performance}
S.~Siami-Namini, N.~Tavakoli, and A.~S. Namin, ``The performance of {LSTM} and
  {BiLSTM} in forecasting time series,'' in \emph{Big Data}.\hskip 1em plus
  0.5em minus 0.4em\relax IEEE, 2019.

\bibitem{agarap2018deep}
A.~F. Agarap, ``Deep learning using rectified linear units (relu),''
  \emph{arXiv preprint arXiv:1803.08375}, 2018.

\bibitem{zhang2019gradient}
J.~Zhang, T.~He, S.~Sra, and A.~Jadbabaie, ``Why gradient clipping accelerates
  training: A theoretical justification for adaptivity,'' \emph{arXiv preprint
  arXiv:1905.11881}, 2019.

\bibitem{mcmahan2017communication}
B.~McMahan, E.~Moore, D.~Ramage, S.~Hampson, and B.~A. y~Arcas,
  ``Communication-efficient learning of deep networks from decentralized
  data,'' in \emph{Artificial intelligence and statistics}.\hskip 1em plus
  0.5em minus 0.4em\relax PMLR, 2017.

\bibitem{nilsson2018performance}
A.~Nilsson, S.~Smith, G.~Ulm, E.~Gustavsson, and M.~Jirstrand, ``A performance
  evaluation of federated learning algorithms,'' in \emph{DIDL}.\hskip 1em plus
  0.5em minus 0.4em\relax ACM, 2018.

\bibitem{lee2007rexnord}
J.~Lee, H.~Qiu, G.~Yu, J.~Lin \emph{et~al.}, ``Rexnord technical services,''
  \emph{Bearing Data Set, IMS, University of Cincinnati, NASA Ames Prognostics
  Data Repository}, 2007.

\bibitem{sakib2020migrating}
S.~Sakib, M.~M. Fouda, Z.~M. Fadlullah, and N.~Nasser, ``Migrating intelligence
  from cloud to ultra-edge smart iot sensor based on deep learning: An
  arrhythmia monitoring use-case,'' in \emph{IWCMC}.\hskip 1em plus 0.5em minus
  0.4em\relax IEEE, 2020.

\bibitem{hahn2021self}
T.~V. Hahn and C.~K. Mechefske, ``Self-supervised learning for tool wear
  monitoring with a disentangled-variational-autoencoder,''
  \emph{Hydromechatronics}, 2021.

\bibitem{mostafavi2021novel}
A.~Mostafavi and A.~Sadighi, ``A novel online machine learning approach for
  real-time condition monitoring of rotating machines,'' in
  \emph{{ICRoM}}.\hskip 1em plus 0.5em minus 0.4em\relax IEEE, 2021.

\bibitem{wang2019edge}
X.~Wang, Y.~Han, C.~Wang, Q.~Zhao, X.~Chen, and M.~Chen, ``In-edge ai:
  Intelligentizing mobile edge computing, caching and communication by
  federated learning,'' \emph{IEEE Network}, vol.~33, no.~5, 2019.

\bibitem{ye2020edgefed}
Y.~Ye, S.~Li, F.~Liu, Y.~Tang, and W.~Hu, ``Edgefed: Optimized federated
  learning based on edge computing,'' \emph{IEEE Access}, vol.~8, 2020.

\bibitem{wu2020personalized}
Q.~Wu, K.~He, and X.~Chen, ``Personalized federated learning for intelligent
  iot applications: A cloud-edge based framework,'' \emph{IEEE Open Journal of
  the Computer Society}, vol.~1, 2020.

\bibitem{dhada2020federated}
M.~Dhada, A.~Parlikad, and A.~S. Palau, ``Federated learning for collaborative
  prognosis,'' \emph{COPEN 2019}, 2020.

\bibitem{zhang2020blockchain}
W.~Zhang, Q.~Lu, Q.~Yu, Z.~Li, Y.~Liu, S.~K. Lo, S.~Chen, X.~Xu, and L.~Zhu,
  ``Blockchain-based federated learning for device failure detection in
  industrial iot,'' \emph{IEEE Internet of Things Journal}, vol.~8, no.~7,
  2020.

\end{thebibliography}
\end{document}